# Further Experimental Evidence against the Utility of Occam's Razor

**Geoffrey I. Webb**                                            WEBB@DEAKIN.EDU.AU
*School of Computing and Mathematics*
*Deakin University*
*Geelong, Vic, 3217, Australia.*

## Abstract

This paper presents new experimental evidence against the utility of Occam's razor. A systematic procedure is presented for post-processing decision trees produced by C4.5. This procedure was derived by rejecting Occam's razor and instead attending to the assumption that similar objects are likely to belong to the same class. It increases a decision tree's complexity without altering the performance of that tree on the training data from which it is inferred. The resulting more complex decision trees are demonstrated to have, on average, for a variety of common learning tasks, higher predictive accuracy than the less complex original decision trees. This result raises considerable doubt about the utility of Occam's razor as it is commonly applied in modern machine learning.

## 1. Introduction

In the fourteenth century William of Occam stated "plurality should not be assumed without necessity". This principle has since become known as Occam's razor. Occam's razor was originally intended as a basis for determining one's ontology. However, in modern times it has been widely reinterpreted and adopted as an epistemological principle—a means of selecting between alternative theories as well as ontologies. Modern reinterpretations of Occam's razor are widely employed in classification learning. However, the utility of this principle has been subject to widespread theoretical and experimental attack. This paper adds to this debate by providing further experimental evidence against the utility of the modern interpretation of Occam's razor. This evidence takes the form of a systematic procedure for adding non-redundant complexity to classifiers in a manner that is demonstrated to frequently improve predictive accuracy.

The modern interpretation of Occam's razor has been characterized as "of two hypotheses H and H', both of which explain E, the simpler is to be preferred" (Good, 1977). However, this does not specify what aspect of a theory should be measured for simplicity. Syntactic, semantic, epistemological and pragmatic simplicity are all alternative criteria that can and have been employed Bunge (1963). In practice, the common use of Occam's razor in machine learning seeks to minimize surface syntactic complexity. It is this interpretation that this paper addresses.

It is to be assumed that Occam's razor is usually applied in the expectation that its application will, in general, lead to some particular form of advantage. There is no widely accepted articulation of precisely how Occam's razor should be applied or what advantages are to be expected from its application in classification learning. However, the literature does contain two statements that seem to capture at least one widely adopted approach to the principle. Blumer, Ehrenfeucht, Haussler, and Warmuth (1987) suggest that to wield Occam's razor is to adopt the goal of discovering "the simplest hypothesis that is consistent with the sample data" with the expectation that the simplest hypothesis will "perform well on further observations taken from the same source". Quinlan (1986) states

> "Given a choice between two decision trees, each of which is correct over the training set, it seems sensible to prefer the simpler one on the grounds that it is more likely to capture structure inherent in the problem. The simpler tree would therefore be expected to classify correctly more objects outside the training set."

While these statements would not necessarily be accepted by all proponents of Occam's razor, they capture the form of Occam's razor that this paper seeks to address—a learning bias toward classifiers that minimize surface syntactic complexity in the expectation of maximizing predictive accuracy.

Both of the above statements of Occam's razor restrict themselves to classifiers that correctly classify all objects in a training set. Many modern machine learning systems incorporate learning biases that tolerate small levels of misclassification of the training data (Clark & Niblett, 1989; Michalski, 1984; Quinlan, 1986, 1990, for example). In this context, and extending the scope of the definition beyond decision trees to classifiers in general, it seems reasonable to modify Quinlan's (1986) statement (above) to

> Given a choice between two plausible classifiers that perform identically on the training set, the simpler classifier is expected to classify correctly more objects outside the training set.

This will be referred to as the *Occam thesis*.

The concept of *identical performance on a training set* could be defined in many different ways. It might be tempting to opt for a definition that requires identical error rates when two classifiers are applied to the training set. A less strict interpretation might allow two classifiers to have differing error rates so long as the difference is within some statistical confidence limit. However, to maximize the applicability of its results, this paper will adopt a very strict interpretation of identical performance—that for every object $o$ in the training set, both classifiers provide the same classification for $o$.

It should be noted that the Occam thesis is not claiming that for any two classifiers with equal empirical support the least complex will always have greater predictive accuracy on previously unseen objects. Rather, it is claiming that more frequently than not the less complex will have higher predictive accuracy.

This paper first examines some arguments for and against the Occam thesis. It then presents new evidence against the thesis. This evidence was acquired by using a learning algorithm that post-processes decision trees learnt by C4.5. This post-processor was developed by rejecting the Occam thesis and instead attending to the assumption that similarity is predictive of class. The post-processor systematically adds complexity to decision trees without altering their performance on the training data. This is demonstrated to lead to an increase in predictive accuracy on previously unseen objects for a range of 'real-world' learning tasks. This evidence is taken as incompatible with the Occam thesis.

## 2. Previous Theoretical and Experimental Work

To provide a context for the new evidence against the Occam thesis, it is worth briefly examining previous relevant theoretical and experimental work. Where relevant, an outline will be provided of reasons why each contribution may have failed to persuade the other side of the debate.

### 2.1 The Law of Conservation of Generalization Performance

The conservation law of generalization performance (Schaffer, 1994) proves that no learning bias can outperform any other bias over the space of all possible learning tasks[1]. It follows that if Occam's razor is a valuable learning bias, it can only be so for some subset of all possible learning tasks. It might be argued that the set of 'real-world' learning tasks is such a subset.

This paper is predicated on accepting the proposition that the set of 'real-world' learning tasks is distinguished from the set of all possible learning tasks in respects that render the conservation law inapplicable. Rao, Gordon, and Spears (1995) argue that this is the case because learning tasks in our universe are not uniformly distributed across the space of all possible learning tasks.

But why should this be so? One argument in support of this proposition is as follows. 'Real-world' learning tasks are defined by people for use with machine learning systems. To this end, the task constructors will have sought to ensure that the independent variables (class attributes) are related to the dependent variables (other attributes) in ways that can be captured within the space of classifiers that are made available for the learning system. Actual machine learning tasks are not drawn randomly from the space of all possible learning tasks. The human involvement in the formulation of the problems ensures this.

As a simple thought experiment in support of this proposition, consider a learning task for which the class attribute is generated by a random number generator and in no way relates to the other attributes. The majority of machine learning researchers would not be in the slightest disconcerted if their systems failed to perform well when trained on such data. As a further example, consider a learning task for which the class attribute is a simple count of the number of missing attribute values for an object. Assume that such a learning task was submitted to a system, such as C4.5 (Quinlan, 1993), that develops classifiers that have no mechanism for testing during classification whether an attribute value is missing. Again, the majority of machine learning researchers would be unconcerned that their systems failed to perform well in such circumstances. Machine learning is simply unsuited to such tasks. A knowledgeable user would not apply machine learning to such data, at least not in the expectation of obtaining a useful classifier therefrom.

This paper explores the applicability of the Occam thesis to 'real-world' learning tasks.

### 2.2 Other Theoretical Objections to the Occam Thesis

Most machine learning systems explicitly or implicitly employ Occam's razor. In addition to its almost universal use in machine learning, the principle of Occam's razor is widely

---

1. The law is only proved for discrete valued learning tasks, but there is no reason to believe it does not also apply to continuous valued tasks accepted in general scientific practice. That this has persisted, despite Occam's razor being subjected to extensive philosophical, theoretical and empirical attack, suggests that these attacks have not been found persuasive.

On the philosophical front, to summarize Bunge (1963), the complexity of a theory (classifier) depends entirely upon the language in which it is encoded. To claim that the acceptability of a theory depends upon the language in which it happens to be expressed appears indefensible. Further, there is no obvious theoretical relationship between syntactic complexity and the quality of a theory, other than the possibility that the world is intrinsically simple and that the use of Occam's razor enables the discovery of that intrinsic simplicity. However, even if the world is intrinsically simple, there is no reason why that simplicity should correspond to syntactic simplicity in an arbitrary language.

To merely state that a less complex explanation is preferable does not specify by what criterion it is preferable. The implicit assumption underlying much machine learning research appears to be that, all other things being equal, less complex classifiers will be, in general, more accurate (Blumer et al., 1987; Quinlan, 1986). It is this Occam thesis that this paper seeks to discredit.

On a straight-forward interpretation, for a syntactic measure to be used to predict expected accuracy appears absurd. If two classifiers have identical meaning (such as IF 20≤AGE≤40 THEN POS and IF 20≤AGE≤30 OR 30≤AGE≤40 THEN POS) then it is not possible for their accuracies to differ, no matter how greatly their complexities differ. This simple example highlights the apparent dominance of semantics over syntax in the determination of predictive accuracy.

## 2.3 Previous Experimental Evidence Against the Occam Thesis

On the empirical front, a number of recent experimental results have appeared to conflict with the Occam thesis. Murphy and Pazzani (1994) demonstrated that for a number of artificial classification learning tasks, the simplest consistent decision trees had lower predictive accuracy than slightly more complex consistent trees. Further experimentation, however, showed that these results were dependent upon the complexity of the target concept. A bias toward simplicity performed well when the target concept was best described by a simple classifier and a bias toward complexity performed well when the target concept was best described by a complex classifier (Murphy, 1995). In addition, the simplest classifiers obtained better than average (over all consistent classifiers) predictive accuracy when the data was augmented with irrelevant attributes or attributes strongly correlated to the target concept, but not required for classification.

Webb (1994) presented results that suggest that for a wide range of learning tasks from the UCI repository of learning tasks (Murphy & Aha, 1993), the relative generality of the classifiers is a better predictor of classification performance than is the relative surface syntactic complexity. However, it could be argued that while these results demonstrate that a strategy of selecting the simplest between any pair of theories will not lead to maximization of predictive accuracy, they do not demonstrate that selecting the simplest of all available theories would fail to maximize predictive accuracy.

Schaffer (1992, 1993) has shown that pruning techniques that reduce complexity while decreasing resubstitution accuracy sometimes increase predictive accuracy and sometimes decrease predictive accuracy of inferred decision trees. However, a proponent of the Occam thesis could explain these results in terms of a positive effect from the application of Occam's razor (the reduction of complexity) being counter-balanced by a negative effect from a reduction of empirical support (resubstitution accuracy).

Holte, Acker, and Porter (1989) have shown that specializing small disjuncts (rules with low empirical support) to exclude areas of the instance space occupied by no training objects frequently decreases the error rate of unseen objects covered by those disjuncts. As this specialization involves increasing complexity, this might be viewed as contrary to the Occam thesis. However, the same research shows that the total error rates for the classifiers in which the disjuncts are embedded increases when those disjuncts are specialized. A proponent of the Occam thesis could thus dismiss the relevance of the former results by arguing that the thesis only applies to complete classifiers and not to elements of those classifiers.

## 2.4 Theoretical and Experimental Support for the Occam Thesis

Against these theoretical and experimental objections to the Occam thesis there exists a body of apparent theoretical and empirical support.

Several attempts have been made to provide theoretical support for the Occam thesis in the machine learning context (Blumer et al., 1987; Pearl, 1978; Fayyad & Irani, 1990). However, these proofs apply equally to any systematic learning bias that favors a small subset of the hypothesis space. Indeed, it has been argued that they equally support a preference for classifiers with high complexity (Schaffer, 1993; Berkman & Sandholm, 1995).

Holte (1993) compared learning very simple classification rules with the use of a sophisticated learner of complex decision trees. He found that, for a number of tasks from the UCI repository of machine learning datasets (Murphy & Aha, 1993), the simple rules achieved accuracies of within a few percentage points of the complex trees. This could be considered as supportive of the Occam thesis. However, in no case did the simple rules outperform the more complex decision trees. Nor was it demonstrated that there did not exist yet another learning bias that consistently outperformed both those studied.

A final argument that might be considered to support the Occam thesis is that the majority of machine learning systems employ some form of Occam's razor and they appear to perform well in practice. However, it has not been demonstrated that even better performance would not be obtained if Occam's razor were abandoned.

## 3. New Experimental Evidence Against the Occam Thesis

The theoretical and experimental objections to the Occam thesis do not appear to have greatly diminished the machine learning community's use of Occam's razor. This paper seeks to support objections to the Occam thesis with robust and general experimental counter-evidence. To this end it presents a systematic procedure for increasing the complexity of inferred decision trees without modifying their performance on the training data. This procedure takes the form of a post-processor for decision trees produced by C4.5 (Quinlan, 1993). The application of this procedure to a range of learning tasks from the UCI repository of learning tasks (Murphy & Aha, 1993) is demonstrated to result, on average, in increased predictive accuracy when the inferred decision trees are applied to previously unseen data.

## 3.1 Theoretical Basis for the Decision Tree Post-processor

The similarity assumption is a common assumption in machine learning—that objects that are similar have high probability of belonging to the same class (Rendell & Seshu, 1990). The techniques to be described rely upon this assumption for their theoretical justification rather than upon the Occam thesis.

Starting from the similarity assumption, machine learning can be viewed as the inference of a suitable similarity metric for a learning task. A decision tree can be viewed as a partitioning of the instance space. Each partition, represented by a leaf, contains the objects that are similar in relevant respects and thus are expected to belong to the same class.

This raises the issue of how similarity should be measured. Instance-based learning methods (Aha, Kibler, & Albert, 1991) tend to map the instance space onto an n-dimensional geometric space and then employ geometric distance measures within that space to measure similarity. Such an approach is problematic on a number of grounds. First, it assumes that the underlying metrics of different attributes are commensurable. How is it possible to determine a priori whether a difference of five years in age signifies a greater or lesser difference in similarity than a difference of one inch in height? Second, it assumes that it is possible to provide a priori definitions of similarity with respect to a single attribute. Can one really make a universal prescription that a value of 16 is always more similar to a value of 2 than to a value of 64? Why should it never be the case that the relevant similarity metric is based on the $\log_2$ of the surface value, in which case 16 would be more similar to 64 than to 2?

If we wish to employ induction to learn classifiers expressed in a particular language then it would appear that we are forced to assume that the language in question in some manner captures a relevant aspect of similarity. Any potential leaf of a decision tree presents a plausible similarity metric (all objects that fall within that leaf are similar in some respect). Empirical evaluation (the performance of that leaf on the training set) can then be used to infer the relevance of that similarity metric to the induction task at hand. If a leaf $l$ covers a large number of objects of class $c$ and few of other classes, then this provides evidence that similarity with respect to the tests that define $l$ is predictive of $c$.

Figure 1 illustrates a simple instance space and the partition that C4.5 (Quinlan, 1993) imposes thereon. Note that C4.5 forms nodes for continuous attributes, such as $A$ and $B$, that consist of a test on a *cut* value $x$. This test takes the form $a \leq x$. With respect to Figure 1 there is one such cut, on value 5 for attribute $A$.

C4.5 infers that the relevant similarity metric relates to attribute $A$ only. The partition (shown by a dashed line) is placed at value 5 for attribute $A$. However, if one does not accept the Occam thesis, but does accept the similarity assumption, there is no reason to believe that the area of the instance space for which $B > 5$ and $A \leq 5$ (lightly shaded in Figure 1) should belong to class $+$ (as determined by C4.5) rather than class $-$.

C4.5 uses the Occam thesis to justify the termination of partitioning of the instance space as soon as the decision tree accounts adequately for the training set. In consequence, large areas of the instance space that are occupied by no objects in the training set may be left within partitions for which the similarity assumption provides little support. For example, with respect to Figure 1, it could be argued that a more relevant similarity metric with respect to the region $A \leq 5$ and $B > 5$ is similarity with respect to $B$. Within the entire instance space, all objects with values of $B > 5$ belong to class $-$. There are five such objects. In contrast, there are only three objects with values of $A \leq 5$ that provide the evidence that objects in this area of the instance space belong to class $+$. Each of these tests represents a plausible similarity metric on the basis of the available evidence. Thus, an object within this region will be similar in a plausible respect to three positive and five negative objects. If objects that are similar in relevant respects have high probability of belonging to the same class, and the only other information available is that it is plausible that an object is similar to three positive and five negative objects, then it would appear more probable that the object is negative than positive.

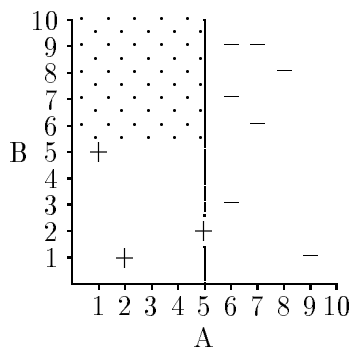

Figure 1: A simple instance space

The disagreement between C4.5 and the similarity assumption in this case contrasts with, for example, the area of the instance space for which $A \leq 5$ and $B < 1$. In this region, the similarity assumption suggests that C4.5's partition is appropriate because all plausible similarity metrics will indicate that an object in this region is similar to positive objects only[2].

The post-processor developed for this research analyses decision trees produced by C4.5 in order to identify such regions—those occupied by no objects from the training set but for which there is evidence (in terms of the similarity assumption) favoring relabeling with a different class to that assigned by C4.5. When such regions are identified, new branches are added to the decision tree, creating new partitionings of the instance space. Both trees must provide identical performance with respect to the training set as only regions of the instance space that are occupied by no objects in the training set are affected.

---

2. To provide an example of an implausible similarity metric, consider the similarity metric defined by the root node, that everything is similar. This will not be plausible as there is too great a level of dissimilarity in classes with respect to this metric. If it were a relevant similarity metric, and the distribution of training examples was representative of the distribution of objects in the domain as a whole, then the similarity assumption would be violated, as similar objects would have probability of just 0.58 of belonging to the same class. This probability can be calculated as follows. The probabilities of an object being $+$ or $-$ are 0.3 and 0.7 respectively. If an object is $+$ then the probability of it belonging to the same class as another object to which it is similar is 0.3. If an object is $-$ then the probability of it belonging to the same class as another object to which it is similar is 0.7. Thus, the probability of an object belonging to the same class as another similar object is $0.3 \times 0.3 + 0.7 \times 0.7 = 0.58$. The numbers involved in this simple example are, of course, too small to reach any such conclusion with a high level of confidence—the example is intended as illustrative only.

It is difficult to see how any plausible metric for complexity could interpret the addition of such branches as not increasing the complexity of the tree.

The end result is that the post-processor adds complexity to the decision tree without altering how the tree applies to the training data. The Occam thesis predicts that this will, in general, lower predictive accuracy while the similarity assumption predicts that it will, in general, increase predictive accuracy. As will be seen, the latter prediction is consistent with experimental evidence and the former is not.

## 3.2 The Post-processor

While the above process could be applied to both continuous and discrete attributes, the current implementation addresses only continuous attributes.

The post-processor operates by examining each leaf $l$ of the tree in turn. For each $l$, each attribute $a$ is considered in turn. For each $a$, all possible thresholds below and above the region of the instance space occupied by objects at $l$ are explored. First, the minimum ($min$) and maximum ($max$) are determined for values of $a$ that are possible for objects that can reach $l$. If $l$ lies below the $\leq$ branch of a split on $a$ then the threshold for that split provides an upper limit ($max$) on values for $a$ at $l$. If it lies below a $>$ branch, the threshold provides a lower limit ($min$). Where the node does not lie below a $\leq$ branch, $max = \infty$. Where the node does not lie below a $>$ branch, $min = -\infty$. Only objects from the training set that have values of $a$ within the range $min..max$ are considered in the following operations.

For each value observed in the training set for the attribute within the allowable range but outside the actual range of values of $a$ for objects at $l$, the evidence is evaluated in support of reclassifying the region above or below that threshold. The level of support for a given threshold is evaluated using a Laplacian accuracy estimate (Niblett & Bratko, 1986). Because each leaf relates to a binary classification (an object belongs to the class in question or does not), the binary form of Laplace is used. For threshold $t$ on attribute $a$ at leaf $l$, the evidence in support of labeling the partition below $t$ with class $n$ is the maximum value for an ancestor node $x$ of $l$ for the formula

$$\frac{P+1}{T+2}$$

where $T$ is the number of objects at $x$ for which $min < a \leq t$; and $P$ is the number of those objects which belong to class $n$.

The evidence in support of labeling a partition above a threshold is calculated identically with the exception that the objects for which $t < a \leq max$ are instead considered.

If the maximum evidence for a new labeling exceeds the evidence for the current labeling of the region, a new branch is added for the appropriate threshold creating a new leaf node labeled with the appropriate class.

In addition to evidence in favor of the current labeling gathered as above, further evidence in support of the current labeling of a region is calculated using the Laplace accuracy estimate considering the objects at the leaf, where $T$ is the number of objects at the leaf and $P$ is the number of those objects that belong to the class with which the node is labeled.

This approach ensures that all new partitions define true regions. That is, for any attribute $a$ and value $v$ it is not possible to partition on $a \leq v$ unless it is possible for both objects from the *domain* with values of $a$ greater than $v$ and objects with values less than or equal to $v$ to reach the node being partitioned (even though no objects from the *training set* will fall within the new partition). In particular, this ensures that the new cuts are not simple duplications of existing cuts at ancestors to the current node. Thus, every modification adds non-redundant complexity to the tree.

This algorithm is presented in Figure 2. It has been implemented as a modification to C4.5 release 6, called C4.5X. The source code for these modifications is available as an on-line appendix to this paper.

In C4.5X, where multiple sets of values equally satisfy the specified constraints and maximize the Laplace function, values of $n_a$ and $n_b$ that are deeper in the tree are selected over those closer to the root and, at a single node, preference for values of $a_a$ and $a_b$ depends upon the order of attributes in the definition of the data and preference for values of $v_a$ and $v_b$ is dependent upon data order. These selection strategies are a side effect of the implementation of the system. There is no reason to believe that the experimental results would differ in general if other strategies were used to select between competing constraints.

By default, C4.5 develops two decision trees each time that it is run, an unpruned and a pruned (simplified) decision tree. C4.5X produces post-processed versions of both of these trees.

## 3.3 Evaluation

To evaluate the post-processor it was applied to all datasets containing continuous attributes from the UCI machine learning repository (Murphy & Aha, 1993) that were then held (due to previous machine learning experimentation) in the local repository at Deakin University. These datasets are believed to be broadly representative of those in the repository as a whole. After experimentation with these eleven data sets, two additional data sets, sick euthyroid and discordant results, were retrieved from the UCI repository and added to the study in order to investigate specific issues, as discussed below.

The resulting thirteen datasets are described in Table 1. The second column contains the number of attributes by which each object is described. Next is the proportion of these that are continuous. The fourth column indicates the proportion of attribute values in the data that are missing (unknown). The fifth column indicates the number of objects that the data set contains. The sixth column indicates the proportion of these that belong to the class represented by the most objects within the data set. The final column indicates the number of classes that the data set describes. Note that the glass type dataset uses the Float/Not Float/Other three class classification rather than the more commonly used six class classification.

Each data set was divided into training and evaluation sets 100 times. Each training set consisted of 80% of the data, randomly selected. Each evaluation set consisted of the remaining 20% of the data. Both C4.5 and C4.5X were applied to each of the resulting 1300 (13 data sets by 100 trials) training and evaluation set pairs.

Let $cases(n)$ denote the set of all training examples that can reach node $n$.

Let $value(a, x)$ denote the value of attribute $a$ for training example $x$.

Let $pos(X, c)$ denote the number of objects of class $c$ in the set of training examples $X$.

Let $Laplace(X, c) = \frac{pos(X,c)+1}{|X|+2}$ where $X$ is a set of training examples, $|X|$ is the number of training examples and $c$ is a class.

Let $upperlim(n, a)$ denote the minimum value of a cut on attribute $a$ for an ancestor node of $n$ for which $n$ lies below a $\leq$ branch. If there is no such cut, $upperlim(n, a) = \infty$. This determines an upper bound on the values for $a$ that may reach $n$.

Let $lowerlim(n, a)$ denote the maximum value of a cut on attribute $a$ for an ancestor node of $n$ for which $n$ lies below a $>$ branch. If there is no such cut, $lowerlim(n, a) = -\infty$. This determines a lower bound on the values for $a$ that may reach $n$.

To post-process leaf $l$ dominated by class $c$

1. Find values of

   $n_a$: $n_a$ is an ancestor of $l$

   $a_a$: $a_a$ is a continuous attribute

   $v_a$: $\exists x: x \in cases(n_a)$ & $v_a = value(a_a, x)$ & $v_a \leq min(v: \exists y: y \in cases(l)$ & $v = value(a_a, y))$ & $v_a > lowerlim(l, a_a)$

   $c_a$: $c_a$ is a class

   that maximize $\mathcal{L}_a = Laplace(\{x: x \in cases(n_a)$ & $value(a_a, x) \leq v_a$ & $value(a_a, x) > lowerlim(l, a_a)\}, c_a)$.

2. Find values of

   $n_b$: $n_b$ is an ancestor of $l$

   $a_b$: $a_b$ is a continuous attribute

   $v_b$: $\exists x: x \in cases(n_b)$ & $v_b = value(a_b, x)$ & $v_b > max(v: \exists y: y \in cases(l)$ & $v = value(a_b, y))$ & $v_b \leq upperlim(l, a_b)$

   $c_b$: $c_b$ is a class

   that maximize $\mathcal{L}_b = Laplace(\{x: x \in cases(n_b)$ & $value(a_b, x) > v_b$ & $value(a_b, x) \leq upperlim(l, a_b)\}, c_b)$.

3. If $\mathcal{L}_a > Laplace(cases(l), c)$ & $\mathcal{L}_a \geq \mathcal{L}_b$ then

   (a) if $c_a \neq c$

      i. replace $l$ with a node $n$ with the test $a_a \leq v_a$.

      ii. set the $\leq$ branch for $n$ to lead to a new leaf for class $c_a$.

      iii. set the $>$ branch for $n$ to lead to $l$.

   else if $\mathcal{L}_b > Laplace(cases(l), c)$

   (b) if $c_b \neq c$

      i. replace $l$ with a node $n$ with the test $a_b \leq v_b$.

      ii. set the $>$ branch for $n$ to lead to a new leaf for class $c_b$.

      iii. set the $\leq$ branch for $n$ to lead to $l$.

Figure 2: C4.5X post-processing algorithm

Table 1: UCI data sets used for experimentation

| Name | No. of Attrs. | % contin­uous | % missing | No. of objects | % most common class | No. of classes |
|---|---|---|---|---|---|---|
| breast cancer Wisconsin | 9 | 100 | <1 | 699 | 66 | 2 |
| Cleveland heart disease | 13 | 46 | <1 | 303 | 54 | 2 |
| credit rating | 15 | 40 | 1 | 690 | 56 | 2 |
| discordant results | 29 | 24 | 6 | 3772 | 98 | 2 |
| echocardiogram | 6 | 83 | 3 | 74 | 68 | 2 |
| glass type | 9 | 100 | 0 | 214 | 40 | 3 |
| hepatitis | 19 | 32 | 6 | 155 | 79 | 2 |
| Hungarian heart disease | 13 | 46 | 20 | 295 | 64 | 2 |
| hypothyroid | 29 | 24 | 6 | 3772 | 92 | 4 |
| iris | 4 | 100 | 0 | 150 | 33 | 3 |
| new thyroid | 5 | 100 | 0 | 215 | 70 | 3 |
| Pima indians diabetes | 8 | 100 | 0 | 768 | 65 | 2 |
| sick euthyroid | 29 | 24 | 6 | 3772 | 94 | 2 |

Table 2 summarizes the percentage predictive accuracy obtained for the unpruned de­cision trees generated by both C4.5 and C4.5X. It presents the mean ($\overline{x}$) and standard deviation ($s$) over each set of 100 trials with respect to each data set for both C4.5 and C4.5X along with the results of a two-tailed matched pairs t-test comparing these means. For twelve of the thirteen data sets C4.5X obtained a higher mean accuracy than C4.5. For the remaining data set, hypothyroid, C4.5 obtained higher mean predictive accuracy than C4.5CS (albeit by a small margin—measured to two decimal places the respective mean ac­curacies were 99.51 and 99.46, respectively). For nine of the data sets the advantage toward C4.5X is statistically significant at the 0.05 level ($p \leq 0.05$), although the advantage with respect to the discordant results data is too small to be apparent when measured to one decimal place (measured to two decimal places the values are 98.58 and 98.62 respectively). The advantage toward C4.5 for the hypothyroid data is also statistically significant at the 0.05 level. The differences in mean predictive accuracy for the Hungarian heart disease, new thyroid and sick euthyroid data sets are not significant at the 0.05 level.

Table 3 uses the same format as Table 2 to summarize the predictive accuracy obtained for the pruned decision trees generated by both C4.5 and C4.5X. For the same twelve data sets C4.5X obtained a higher mean predictive accuracy than C4.5. For the remaining data set, hypothyroid, C4.5 again obtained higher mean predictive accuracy, although again the magnitude of the difference is so small that it is not apparent at the level of precision displayed (measured to two decimal places the mean accuracies are 99.51 and 99.46). For six of the data sets the advantage toward C4.5X is statistically significant at the 0.05 level, although the difference is only apparent at a precision of two decimal places for the discordant results data (99.81 and 99.82, respectively). The advantage toward C4.5 for the hypothyroid data is also statistically significant at the 0.05 level. The differences for breast cancer Wisconsin, echocardiogram, Hungarian heart disease, iris, new thyroid and sick euthyroid are not statistically significant at the 0.05 level.

Table 2: Percentage predictive accuracy for unpruned decision trees.

| Name | C4.5 | | C4.5X | | | |
|------|------|---|-------|---|---|---|
| | $\overline{x}$ | $s$ | $\overline{x}$ | $s$ | $t$ | $p$ |
| breast cancer Wisconsin | 94.1 | 1.8 | 94.4 | 1.7 | −3.2 | 0.002 |
| Cleveland heart disease | 72.8 | 5.0 | 74.4 | 4.8 | −6.1 | 0.000 |
| credit rating | 82.2 | 3.4 | 83.0 | 3.3 | −7.6 | 0.000 |
| discordant results | 98.6 | 0.5 | 98.6 | 0.5 | −5.4 | 0.000 |
| echocardiogram | 72.0 | 9.8 | 73.5 | 10.2 | −2.8 | 0.007 |
| glass type | 74.0 | 7.0 | 75.3 | 7.2 | −4.2 | 0.000 |
| hepatitis | 79.6 | 7.1 | 80.8 | 6.9 | −3.3 | 0.001 |
| Hungarian heart disease | 77.0 | 5.3 | 77.4 | 5.2 | −1.8 | 0.082 |
| hypothyroid | 99.5 | 0.2 | 99.5 | 0.2 | 4.4 | 0.000 |
| iris | 95.4 | 3.4 | 95.7 | 3.5 | −2.2 | 0.028 |
| new thyroid | 89.9 | 4.2 | 90.1 | 4.3 | −1.0 | 0.302 |
| Pima indians diabetes | 70.2 | 3.5 | 71.3 | 3.6 | −8.1 | 0.000 |
| sick euthyroid | 98.7 | 0.5 | 98.7 | 0.5 | −0.0 | 0.963 |

Table 3: Percentage accuracy for pruned decision trees.

| Name | C4.5 | | C4.5X | | | |
|------|------|---|-------|---|---|---|
| | $\overline{x}$ | $s$ | $\overline{x}$ | $s$ | $t$ | $p$ |
| breast cancer Wisconsin | 95.1 | 1.7 | 95.2 | 1.7 | −2.0 | 0.051 |
| Cleveland heart disease | 74.1 | 5.3 | 74.8 | 5.3 | −3.7 | 0.000 |
| credit rating | 84.1 | 3.2 | 84.6 | 3.2 | −5.3 | 0.000 |
| discordant results | 98.8 | 0.4 | 98.8 | 0.4 | −2.6 | 0.010 |
| echocardiogram | 74.2 | 9.3 | 75.1 | 9.8 | −1.6 | 0.1180 |
| glass type | 74.4 | 6.9 | 75.4 | 6.9 | −3.3 | 0.001 |
| hepatitis | 79.9 | 6.2 | 80.7 | 6.2 | −3.0 | 0.003 |
| Hungarian heart disease | 79.2 | 4.9 | 79.4 | 4.8 | −1.0 | 0.310 |
| hypothyroid | 99.5 | 0.2 | 99.5 | 0.2 | 5.4 | 0.000 |
| iris | 95.4 | 3.6 | 95.7 | 3.7 | −1.6 | 0.109 |
| new thyroid | 89.6 | 4.2 | 89.8 | 4.2 | −0.8 | 0.451 |
| Pima indians diabetes | 72.2 | 3.5 | 72.8 | 3.5 | −5.9 | 0.000 |
| sick euthyroid | 98.7 | 0.4 | 98.7 | 0.4 | −0.7 | 0.480 |

After completing experimentation on the initial eleven data sets, the results for the hypothyroid data stood out in stark contrast from those for the other ten. This raised the possibility that there might be distinguishing features of the hypothyroid data that accounted for this difference in performance. Table 1 indicates this data set is clearly distinguishable from the other ten initial data sets in the following six respects—

- having more attributes;

- containing a greater proportion of discrete attributes (which are not directly addressed by C4.5X);

- containing more objects;

- having a greater proportion of the objects belong to the most common class;

- having more classes; and

- producing decision trees of extremely high predictive accuracy without post-processing.

To explore these issues the discordant results and sick euthyroid data sets were retrieved from the UCI repository and added to the study. These data sets are identical to the hypothyroid data set with the exception that each has a different class attribute. All three data sets contain the same objects, described by the same attributes. The addition of the discordant results and sick euthyroid data did little to illuminate this issue however. For all three data sets the changes in accuracy are of very small magnitude. For hypothyroid there is a significant advantage to C4.5. For sick euthyroid there is no significant advantage to either system. For the discordant results data there is a significant advantage to C4.5X.

The question of whether there is a distinguishing feature of the hypothyroid data that explains the observed results remains unanswered. Further investigation of this issue lies beyond the scope of the current paper but remains an interesting direction for future research.

These results suggest that C4.5X's post-processing more frequently increases predictive accuracy than not for the type of data to be found in the UCI repository. (Of the twenty-six comparisons, there was a significant increase for fifteen and there was a significant decrease for only two. A sign test reveals that this rate of success is significant at the 0.05 level, $p = 0.001$.)

Tables 4 and 5 summarize the number of nodes in the decision trees developed. Table 4 addresses unpruned decision trees and Table 5 addresses pruned decision trees. Each post-processing modification replaces a single leaf with a split and two leaves. At most one such modification can be performed per leaf in the original tree. For all data sets the post-processed decision trees are significantly more complex than the original decision trees. In most cases post-processing has increased the mean number of nodes in the decision trees by approximately 50%. This demonstrates that the post-processing is causing substantial change.

## 4. Discussion

The primary objective of this research has been to discredit the Occam thesis. To this end it uses a post-processor that disregards the Occam thesis and instead is theoretically founded upon the similarity assumption. Experimentation with this post-processor has demonstrated that it is possible to develop systematic procedures that, for a range of 'real-world' learning tasks increase the predictive accuracy of inferred decision trees as a result of changes that substantially increase their complexity without altering their performance upon the training data.

Table 4: Number of nodes for unpruned decision trees.

| Name | C4.5 | | C4.5X | | | |
|---|---|---|---|---|---|---|
| | $\overline{x}$ | $s$ | $\overline{x}$ | $s$ | $t$ | $p$ |
| breast cancer Wisconsin | 38.1 | 6.0 | 64.0 | 10.3 | −51.5 | 0.000 |
| Cleveland heart disease | 66.7 | 7.1 | 100.2 | 11.3 | −61.9 | 0.000 |
| credit rating | 117.6 | 18.1 | 177.9 | 28.4 | −44.2 | 0.000 |
| discordant results | 64.0 | 10.6 | 85.2 | 16.2 | −33.3 | 0.000 |
| echocardiogram | 15.4 | 4.1 | 22.1 | 6.3 | −26.1 | 0.000 |
| glass type | 43.0 | 5.2 | 69.7 | 8.4 | −57.2 | 0.000 |
| hepatitis | 24.5 | 4.2 | 34.8 | 6.0 | −49.1 | 0.000 |
| Hungarian heart disease | 62.1 | 7.5 | 94.8 | 13.0 | −50.1 | 0.000 |
| hypothyroid | 29.4 | 4.4 | 47.5 | 7.1 | −57.8 | 0.000 |
| iris | 9.0 | 1.9 | 16.0 | 4.0 | −31.5 | 0.000 |
| new thyroid | 14.7 | 2.4 | 23.4 | 3.8 | −41.5 | 0.000 |
| Pima indians diabetes | 164.8 | 10.8 | 238.8 | 16.3 | −108.9 | 0.000 |
| sick euthyroid | 71.7 | 6.6 | 111.4 | 12.1 | −65.8 | 0.000 |

Table 5: Number of nodes for pruned decision trees.

| Name | C4.5 | | C4.5X | | | |
|---|---|---|---|---|---|---|
| | $\overline{x}$ | $s$ | $\overline{x}$ | $s$ | $t$ | $p$ |
| breast cancer Wisconsin | 19.2 | 5.0 | 33.1 | 8.6 | −34.9 | 0.000 |
| Cleveland heart disease | 44.6 | 8.3 | 68.3 | 12.8 | −43.6 | 0.000 |
| credit rating | 51.2 | 14.8 | 78.4 | 24.2 | −25.8 | 0.000 |
| discordant results | 24.9 | 5.6 | 32.5 | 8.8 | −21.1 | 0.000 |
| echocardiogram | 10.4 | 3.0 | 14.8 | 4.8 | −21.0 | 0.000 |
| glass type | 36.6 | 5.5 | 61.0 | 9.5 | −48.5 | 0.000 |
| hepatitis | 13.7 | 4.8 | 19.8 | 6.6 | −30.7 | 0.000 |
| Hungarian heart disease | 26.8 | 11.4 | 41.2 | 17.3 | −22.1 | 0.000 |
| hypothyroid | 23.6 | 2.9 | 37.1 | 5.6 | −46.7 | 0.000 |
| iris | 8.2 | 1.9 | 14.8 | 3.9 | −30.3 | 0.000 |
| new thyroid | 14.1 | 2.7 | 22.5 | 4.3 | −36.9 | 0.000 |
| Pima indians diabetes | 112.0 | 16.4 | 163.9 | 24.0 | −62.5 | 0.000 |
| sick euthyroid | 46.5 | 5.8 | 72.6 | 8.7 | −76.7 | 0.000 |

It is, in general, difficult to attack the Occam thesis due to the absence of a widely agreed formulation thereof. However, it is far from apparent how the Occam thesis might be recast to both accommodate these experimental results and provide a practical learning bias.

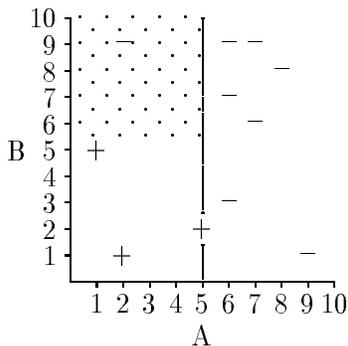

Figure 3: Modified simple instance space

## 4.1 Directions for Future Research

The implications of this research reach beyond its relevance to Occam's razor. The post-processor appears to have practical utility in increasing the quality of inferred decision trees. However, if the objective of the research were to improve predictive accuracy rather than to discredit the Occam thesis, the post-processor would be modified in a number of ways.

The first modification would be to enable the addition of multiple partitions at a single leaf from the original tree. C4.5X selects only the single modification for which there is the maximum support. This design decision originated from a desire to minimize the likelihood of performing modifications that will decrease accuracy. In principle, however, it would appear desirable to select all modifications for which there is strong support, each of which could then be inserted into the tree in order of level of supporting evidence.

Even greater increases in accuracy might be expected if one removed the constraint that the post-processing should not alter the performance of the decision tree with respect to the training set. In this case, new partitions may well be found that employ objects from other regions of the instance space to provide evidence in support of adding partitions that correct misclassifications of small numbers of objects at a leaf node from the original tree. The similarity assumption would provide strong evidence for such repartitioning. Such a situation would occur, for example, with respect to the learning problem illustrated in Figure 1, if there was an additional object of class − with attribute values A=2 and B=9. This is illustrated in Figure 3. In this case C4.5 would still create the indicated partitions. However, C4.5X would be unable to relabel the area containing the additional object due to the constraint that it not alter the performance of the original decision tree with respect to the training set. Thus the addition of the object prevents C4.5X from relabeling the shaded region even though, on the basis of the similarity assumption, it improves the evidence in support of that relabeling.

Such an extended post-processor would encourage the following model of inductive inference of decision trees. The role of C4.5 (or a similar system) would be to identify clusters of objects within the instance space that should be grouped under a single leaf node. A second stage would then analyze regions of the instance space that lie outside those clusters in order to allocate classes to those regions. Current decision tree learners, motivated by the Occam thesis, ignore this second stage, leaving regions outside the identified clusters associated with whatever classes have been assigned to them as a by-product of the cluster identification process.

## 4.2 Other Related Research

A number of researchers have developed learning systems that can be viewed as considering evidence from neighboring regions of the instance space in order to derive classifications within regions of the instance space that are not occupied by examples from the training set. Ting (1994) does this explicitly, by examining the training set to directly explore the neighborhood of the object to be classified. This system uses instance based learning for classification within nodes of a decision tree with low empirical support (small disjuncts).

A number of other systems can also be viewed as considering evidence from neighboring regions for classification. These systems learn and then apply multiple classifiers (Ali, Brunk, & Pazzani, 1994; Nock & Gascuel, 1995; Oliver & Hand, 1995). In such a context, any point within a region of the instance space that is occupied by no training objects is likely to be covered by multiple leaves or rules. Of these, the leaf or rule with the greatest empirical support will be used for classification.

C4.5X uses two distinct criteria for evaluating potential splits. The standard C4.5 stage of tree induction employs an information measure to select splits. The post-processor uses a Laplace accuracy estimate. Similar uses of dual criteria have been investigated elsewhere. Quinlan (1991) employs a Laplace accuracy estimate considering neighboring regions of the instance space to estimate the accuracy of small disjuncts. Lubinsky (1995) and Brodley (1995) employ resubstitution accuracy to select splits near the leaves during induction of decision trees.

By adding a split to a leaf, C4.5X is specializing with respect to the class at that leaf (and generalizing with respect to the class of the new leaf). Holte et al. (1989) explored a number of techniques for specializing small disjuncts. C4.5X differs in that all leaves are candidates for specialization, not just those with low empirical support. It further differs in the manner in which it selects the specialization to perform by considering the evidence in support of alternative splits rather than just the strength of the evidence in support of individual potential conditions for the current disjunct.

## 4.3 Bias Versus Variance

Breiman, Friedman, Olshen, and Stone (1984) provide an analysis of complexity and induction in terms of a trade-off between *bias* and *variance*. A classifier partitions the instance space into regions. When these regions are too large, the degree of fit to an accurate partitioning of the instance space will be poor, increasing error rates. This effect is called *bias*. When the regions are too small, the probability that individual regions are labeled with the wrong class is increased. This effect, called *variance*, also increases error rates. According to this analysis, due to variance, too fine a partitioning of the instance space tends to increase the error rate while, due to bias, too coarse a partitioning also tends to increase the error rate.

Increasing the complexity of a decision tree creates finer partitionings of the instance space. This analysis can be used to argue against the addition of undue complexity to decision trees on the ground that it will increase variance and hence the error rate.

However, the success of C4.5X in decreasing the error rate demonstrates that it is successfully managing the bias/variance trade-off when it introduces complexity to the decision tree. By using evidence from neighboring regions of the instance space, C4.5X is successful in increasing the error rate resulting from variance at a lower rate than it decreases the error rate resulting from bias. The success of C4.5X demonstrates that it is not adding *undue* complexity to C4.5's decision trees.

## 4.4 Minimum Encoding Length Induction

Minimum encoding length approaches perform induction by seeking a theory that enables the most compact encoding of both the theory and available data. Two key approaches have been developed, Minimum Message Length (MML) (Wallace & Boulton, 1968) and Minimum Description Length (MDL) (Rissanen, 1983). Both approaches admit to probabilistic interpretations. Given prior probabilities for both theories and data, minimization of the MML encoding closely approximates maximization of posterior probability (Wallace & Freeman, 1987). An MDL code length defines an upper bound on "unconditional likelihood" (Rissanen, 1987).

The two approaches differ in that MDL employs a universal prior, which Rissanen (1983) explicitly justifies in terms of Occam's razor, while MML allows the specification of distinct appropriate priors for each induction task. However, in practice, a default prior is usually employed for MML, one that appears to also derive its justification from Occam's razor.

Neither MDL nor MML with its default prior would add complexity to a decision tree if doing so were justified solely on the basis of evidence from neighboring regions of the instance space. The evidence from the study presented herein appears to support the potential desirability of doing so. This casts some doubt upon the utility of the universal prior employed by MDL and the default prior usually employed with MML, at least with respect to their use for maximizing predictive accuracy.

It should be noted, however, that the probabilistic interpretation of these minimum encoding length techniques indicates that encoding length minimization represents maximization of posterior probability or of unconditional likelihood. Maximization of these factors is not necessarily directly linked with maximizing predictive accuracy.

## 4.5 Appropriate Application of Grafting and Pruning

It is important to note that although this paper calls into question the value of learning biases that penalize complexity, in no way does it provide support for learning biases that encourage complexity for its own sake. C4.5X only grafts new nodes onto a decision tree when there is empirical support for doing so.

Nor do the results in any way argue against the appropriate use of decision tree pruning. To generate its pruned trees, C4.5 removes branches where statistical estimates of the upper bounds on the error rates indicate that these will not increase if the branch is removed. It could be argued that C4.5 only reduces complexity when there is empirical support for doing so. It is interesting to note that for eight of the thirteen data sets examined, C4.5X's post-processing of the pruned trees resulted in higher average predictive accuracy than post-processing of unpruned trees. These results suggest that both pruning and grafting can play a valuable role when applied appropriately.

## 5. Conclusion

This paper presents a systematic procedure for adding complexity to inferred decision trees without altering their performance on the training data. This procedure has been demonstrated to lead to increases in predictive accuracy for a range of learning tasks when applied to both pruned and unpruned trees inferred by C4.5. For only one of the thirteen learning tasks examined did the procedure lead to a statistically significant loss in accuracy and in this case the magnitude of the difference in mean accuracy was extremely small. On the face of it, this provides strong experimental evidence against the Occam thesis.

This post-processing technique was developed by rejecting the Occam thesis and instead attending to the similarity assumption—that similar objects have high probability of belonging to the same class.

The procedure developed was constrained by the need to ensure that the revised decision tree performed identically to the original decision tree with respect to the training data. This constraint arose from the desire to obtain experimental evidence against the Occam thesis. It is possible that if this constraint is removed, the basic techniques outlined in this paper could result in even greater improvements in predictive accuracy than those reported herein.

This research has considered only one version of Occam's razor that favors minimization of syntactic complexity in the expectation that this will tend to increase predictive accuracy. Other interpretations of Occam's razor are also possible, such as that one should minimize semantic complexity. While others (Bunge, 1963) have provided philosophical objections to such formulations of Occam's razor, this paper has not sought to investigate them.

The version of Occam's razor examined in this research has been used widely in machine learning with apparent success. The objections to this principle that have been substantiated by this research raise the question, why has it had such apparent success if it is so flawed? Webb (1994) suggests that the apparent success of the principle has been due to the manner in which syntactic complexity is usually associated with other relevant qualities of inferred classifiers such as generality or prior probability. If this thesis is accepted then one of the key challenges facing machine learning is to understand these deeper qualities and to employ that understanding to place machine learning on a sounder theoretical footing. This paper offers a small contribution in this direction by demonstrating that minimization of surface syntactic complexity does not, in itself, in general maximize the predictive accuracy of inferred classifiers.

It is nonetheless important to realize that, the thrust of this paper notwithstanding, Occam's razor will often be a useful learning bias to employ. This is because there will frequently be good pragmatic reasons for preferring a simple hypothesis. A simple hypothesis will in general be easier to understand, communicate and employ. A preference for simple hypotheses cannot be justified in terms of expected predictive accuracy but may be justified on pragmatic grounds.

## Acknowledgements

This research has been supported by the Australian Research Council. I am grateful to Charlie Clelland, David Dowe, Doug Newlands, Ross Quinlan and anonymous reviewers for extremely valuable comments from which the paper has benefited greatly.